\documentclass[a4paper,twoside]{article}

\usepackage{epsfig}
\usepackage{subcaption}
\usepackage{calc}
\usepackage{amssymb}
\usepackage{amstext}
\usepackage{amsmath}
\usepackage{amsthm}
\usepackage{multicol}
\usepackage{pslatex}
\usepackage{apalike}
\usepackage[bottom]{footmisc}
\usepackage{booktabs}

\usepackage{todonotes}
    \presetkeys{todonotes}{inline}{}
\usepackage{xcolor}
\usepackage{acro}
\usepackage{hyperref}
\usepackage[capitalise, noabbrev]{cleveref}
\usepackage[natbib,style=iso-authoryear]{biblatex}
\usepackage{tabularx}
\usepackage{enumitem}
\usepackage{amsthm}
\usepackage{algorithm}
\usepackage[noend]{algpseudocode}
    \hypersetup{unicode}
    \hypersetup{breaklinks=true}
\usepackage{csquotes}
\usepackage{soul}

\usepackage{SCITEPRESS}     

\DeclareAcronym{aps}{
    short=APS,
    long=Advanced Planning and Scheduling,
}
\DeclareAcronym{rcpsp}{
    short=RCPSP,
    long=Resource-Constrained Project Scheduling Problem
}
\DeclareAcronym{ebm}{
    short=EBM,
    long=Execution Bottleneck Machine
}

\DeclareAcronym{docplex}{
    short=DOcplex,
    long=IBM Decision Optimization CPLEX
}

\DeclareAcronym{cpopt}{
    short=CP Optimizer,
    long=IBM ILOG Constraint-Programming Optimizer
}

\DeclareAcronym{iira}{
    short=IIRA,
    long=Identification Indicator-based Relaxing Algorithm
}
\DeclareAcronym{ssira}{
    short=SSIRA,
    long=Schedule Suffix Interval Relaxing Algorithm
}

\DeclareAcronym{mw}{
    short=MW,
    long=Machine Workload
}
\DeclareAcronym{mur}{
    short=MUR,
    long=Machine Utilization Rate
}
\DeclareAcronym{auad}{
    short=AUAD,
    long=Average Uninterrupted Active Duration
}
\DeclareAcronym{rs}{
    short=RS,
    long=Resource Strength
}
\DeclareAcronym{rc}{
    short=RC,
    long=Resource Constrainedness
}
\DeclareAcronym{ql}{
    short=QL,
    long=Queue Length
}

\DeclareAcronym{mrw}{
    short=MRW,
    long=Machine Resource Workload
}
\DeclareAcronym{mrur}{
    short=MRUR,
    long=Machine Resource Utilization Rate
}
\DeclareAcronym{auau}{
    short=AUAU,
    long=Average Uninterrupted Active Utilization
}
\DeclareAcronym{pru}{
    short=PRU,
    long=Period Resource Utilization
}

\newcommand{\N}{\mathbb{N}}
\newcommand{\Nzero}{\mathbb{N}_0}

\theoremstyle{plain}

\newtheorem{defn}{Definition}

\theoremstyle{remark}

\newcommand{\capsection}[1]{\section{\uppercase{#1}}}

\newlist{steps}{enumerate}{10}
\setlist[steps]{label*=\arabic*.}    
\crefname{stepsi}{step}{steps}
\Crefname{stepsi}{Step}{Steps}

\newlist{conditions}{enumerate}{10}
\setlist[conditions]{label=\roman*)}
\crefname{conditionsi}{condition}{conditions}
\Crefname{conditionsi}{Condition}{Conditions}

\algrenewcommand\algorithmicdo{\textbf{:}}
\algrenewcommand\algorithmicthen{\textbf{:}}
\algrenewcommand\algorithmicelse{\textbf{else:}}

\algnewcommand\algorithmicinput{\textbf{Input:}}
\algnewcommand\algorithmicinputc{\phantom{\textbf{Input:}}}
\algnewcommand\algorithmicoutput{\textbf{Output:}}
\algnewcommand\algorithmicoutputc{\phantom{\textbf{Output:}}}
\algnewcommand\algorithmicparams{\textbf{Parameters:}}
\algnewcommand\algorithmicparamsc{\phantom{\textbf{Parameters:}}}
\algnewcommand\algorithmicforiter{\phantom{\textbf{For}}}

\algnewcommand\Input{\item[\algorithmicinput]}%
\algnewcommand\Inputc{\item[\algorithmicinputc]}%
\algnewcommand\Output{\item[\algorithmicoutput]}%
\algnewcommand\Outputc{\item[\algorithmicoutputc]}%
\algnewcommand\Params{\item[\algorithmicparams]}%
\algnewcommand\Paramsc{\item[\algorithmicparamsc]}%
\algnewcommand\Note{\footnotesize\item[\textit{Note:}]}%
\algnewcommand\Notec{\footnotesize\item[\phantom{\textit{Note:}}]}%
\newcommand{\Callref}[3]{\hyperref[{#3}]{\Call{#1}{#2}}}
\newcommand{\algnameref}[2]{\hyperref[{#2}]{\textproc{#1}}}

\algdef{SE}[REPEATITER]{Repeat}{ForIter}{\textbf{repeat:}}[1]{\textbf{for}\ #1\ \textbf{iterations}}%

\newcounter{algcounter}

\algrenewcommand\algorithmicindent{1.0em}%

\newcommand{\defeq}{\overset{\text{def}}{=}}
\newcommand{\intinterval}[2]{\{{#1}, \dots, {#2}\}}

\DeclareMathOperator*{\argmax}{\operatorname{argmax}}

\newcommand{\abs}[1]{\left|{#1}\right|}

\NewDocumentCommand{\supscriptable}{m m e{^}}{
    \IfValueTF{#3}{%
        #1^{#3#2}%
    }{%
        #1^{#2}%
    }
}

\newcommand{\Problem}{$PSm \;|\; intree \;|\; \sum_{j} w_j T_j$}

\newcommand{\Instance}{\mathcal{I}}
\newcommand{\precedence}[2]{#1 \hspace{-0.2em} \rightarrow \hspace{-0.2em} #2}
\newcommand{\Jobs}{\mathcal{J}}
\newcommand{\Precedences}{\mathcal{P}}
\newcommand{\Projects}{\mathrm{P}}
\newcommand{\Resources}{\mathcal{R}}
\newcommand{\horizon}{\mathcal{T}}

\newcommand{\duration}[1]{p_{#1}}
\newcommand{\deadline}[1]{d_{#1}}
\newcommand{\tardinessweight}[1]{w_{#1}}
\newcommand{\capacity}[2]{\supscriptable{R_{#1}}{(#2)}}
\newcommand{\capacityf}[1]{R_{#1}}

\newcommand{\consumption}[2]{r_{#1 #2}}

\newcommand{\jobstart}[1]{S_{#1}}
\newcommand{\Schedule}{S}
\newcommand{\jobend}[1]{C_{#1}}
\newcommand{\Completions}{C}
\newcommand{\tardiness}[1]{T_{#1}}

\newcommand{\targetProject}{p}
\newcommand{\closure}[1]{\mathcal{L}(#1)}

\newcommand{\migration}[5]{(#1, #2, #3, #4, #5)}
\newcommand{\Migrations}{\mathcal{M}}
\newcommand{\addition}[4]{(#1, #2, #3, #4)}
\newcommand{\Additions}{\mathcal{A}}

\newcommand{\modelConsumption}[3]{\operatorname{c}_{#1 #2}^{(#3)}}

\newcommand{\resourceLoad}[1]{L_{#1}}

\newcommand{\indicator}[2]{\operatorname{#1}_{#2}}

\newcommand{\indMRUR}[1]{\indicator{MRUR}{#1}}
\newcommand{\indAUAU}[1]{\indicator{AUAU}{#1}}
\newcommand{\indPRU}[2]{\indicator{PRU}{#1}^{(#2)}}

\newcommand{\JobsOnResource}[1]{\Jobs_{#1}}

\newcommand{\JobsOnResourceInPeriod}[2]{\Jobs_{#1}^{UAP(#2)}}

\newcommand{\algIndicator}{\mathrm{I}}
\newcommand{\algConvolution}{\mathrm{C}}
\newcommand{\algMaxiter}{I_{\max}}
\newcommand{\algGranularity}{\mathrm{G}}
\newcommand{\algMaxperiods}{P_{\max}}
\newcommand{\algImprovement}{\Delta}
\newcommand{\algMaxintervals}{IT_{\max}}
\newcommand{\algSortkey}{\mathcal{K}}

\newcommand{\algSortkeyTime}{\algSortkey_{t}}
\newcommand{\algSortkeyImprovement}{\algSortkey_{\Delta\jobstart{}}}

\newcommand{\kpiImprovement}{\Delta\tardiness{\targetProject}}

\newcommand{\kpiDiff}{\Delta\jobstart{}}

\newcommand{\inst}[2]{\texttt{instance0#1\ifthenelse{\equal{#2}{}}{}{\_#2}}}

\newcommand{\relaxedSchedule}[1]{\sigma^{(#1)}}
\newcommand{\relaxedjobstart}[2]{\relaxedSchedule{#1}_{#2}}

\begin{document}

\title{Bottleneck Identification in Resource-Constrained Project Scheduling via Constraint Relaxation}

\author{%
\authorname{Lukáš Nedbálek\sup{1,2} and %
Antonín Novák\sup{2}}
\affiliation{\sup{1}Faculty of Mathematics and Physics, Charles University, Prague, CZ}
\affiliation{\sup{2}Czech Institute of Informatics, Robotics and Cybernetics, Czech Technical University in Prague, CZ}
\email{\href{mailto:lukas.nedbalek@email.cz}{lukas.nedbalek@email.cz}, \href{mailto:antonin.novak@cvut.cz}{antonin.novak@cvut.cz}}
}


\abstract{
In realistic production scenarios, Advanced Planning and Scheduling (APS) tools often require manual intervention by production planners, as the system works with incomplete information, resulting in suboptimal schedules.
Often, the preferable solution is not found just because of the too-restrictive constraints specifying the optimization problem, representing bottlenecks in the schedule.
To provide computer-assisted support for decision-making, we aim to automatically identify bottlenecks in the given schedule while linking them to the particular constraints to be relaxed.
In this work, we address the problem
of reducing the tardiness of a particular project in an obtained schedule
in the resource-constrained project scheduling problem by relaxing constraints related to identified bottlenecks.
We develop two methods for this purpose.
The first method adapts existing approaches from the job shop literature
and utilizes them for so-called untargeted relaxations.
The second method identifies potential improvements in relaxed versions of the problem
and proposes targeted relaxations.
Surprisingly, the untargeted relaxations result in improvements comparable to the targeted relaxations.
}

\onecolumn \maketitle \normalsize \setcounter{footnote}{0} \vfill

\capsection{Introduction}

In the modern manufacturing industry, \ac{aps} tools are used to schedule production automatically.
However, not all parameters and information are available to the \ac{aps} systems in practice.
Thus, the solutions obtained with the scheduling tools may not be preferable to the users.
This leads to repeated interactions of the production planners with the \ac{aps} system, adjusting the problem parameters to obtain an acceptable schedule.

\begin{figure}[ht]
    \centering
    \includegraphics[width=0.97\linewidth]{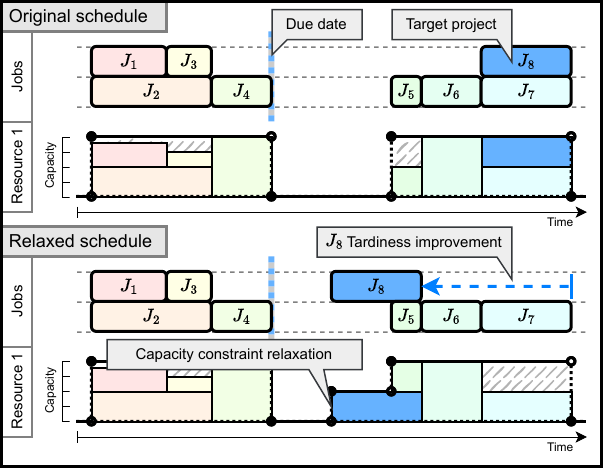}
    \caption{
        Example of an original and relaxed schedule with 8 jobs and a single resource.
        The \emph{Jobs} segments show the (overlapping) scheduling of jobs in time.
        The \emph{Resource~1} segments show the cumulative consumption of the single resource by the scheduled jobs.
        In the \emph{Relaxed schedule} the \emph{Capacity constraint relaxation} refers to the temporary increase of the resource capacity.
    }\label{fig:illustrative_example}
\end{figure}

We address a common problem in production---reducing the delay of a selected project in an existing production schedule.
We achieve this by identifying \textit{manufacturing bottlenecks} in the schedule and related constraints.
Bottlenecks represent resources or constraints with the most significant impact on production performance.
The algorithm identifies these bottlenecks and suggests appropriate relaxations in the form of a modified schedule.
The modified schedule is then offered to the decision maker to compare the improvements,
and the changes can then be accepted, rejected, or augmented to suit the current needs.

See an example in Figure~\ref{fig:illustrative_example}.
In the original schedule,
the target project $J_8$ should preferably have been completed earlier,
but the resource capacity is insufficient.
The target project could be completed earlier by relaxing the capacity constraint,
as shown in the relaxed schedule.
This relaxation can then be interpreted as, e.g., adjusting the start of a shift for a single employee.

The above example is a particularly simple case of the problem;
in reality, activities of different projects have complex interactions across multiple resources,
making the identification of the constraints to be relaxed a difficult problem.
In some sense, this approach is similar to the concept of duality known in continuous optimization
(e.g., shadow prices in Linear Programming), which does not apply to discrete optimization problems.

We develop two different methods to address the problem.
The first method adapts existing approaches from the literature,
which address the problem in simpler scheduling models.
The second method utilizes relaxations that focus specifically on the target project.
To compare the methods, we present a set of problem instances designed to model the addressed scheduling problem, and we evaluate the two methods using the presented problem instances.

\capsection{Related Work} \label{sec:related-works}
Following the work of \citet{Wang2016},
we focus on \emph{Execution bottleneck machines},
which are bottlenecks in a constructed schedule for a given problem instance, i.e., not for the model of the production but for the specific data.
The identified execution bottlenecks may vary between problem instances and their constructed schedules.
Focusing on execution bottlenecks aims to improve performance for the specific problem instance, i.e., \textit{case-based relaxation}.

\subsection{Bottlenecks in the \acs{rcpsp}} \label{subsec:related-works/bottlenecks-in-scheduling/bottlenecks-in-the-rcpsp}

To the best of our knowledge,
no relevant research focuses mainly on identifying bottlenecks in the \ac{rcpsp}.
The closest research on bottlenecks can be found for the Job-Shop problem---scheduling on unit-capacity resources.

\citet{Luo2023} studied how identifying bottleneck machines can guide the scheduling
process of a genetic algorithm.
In their case study, \citet{Arkhipov2017} proposed a heuristic approach
for estimating project makespan and resource load profiles.
Those estimations are, in turn, used to identify bottleneck resources.
However, the identified bottleneck resources were not addressed further.

\subsection{Relaxing the Identified Bottlenecks} \label{subsec:related-works/bottlenecks-in-scheduling/relaxing-bottlenecks}

\citet{Lawrence1994} studied how identified bottlenecks shift between machines
in response to introducing relaxations to the original problem.
They relaxed \enquote{short-run} bottlenecks by increasing the capacity of the identified bottleneck resources
and then observed whether the bottlenecks shifted to a different machine.
The authors observed that while such relaxations are effective at relaxing local bottlenecks,
they also increase the \enquote{bottleneck shiftiness}.

In their study, \citet{Zhang2012} addressed the Job-Shop problem by first relaxing its capacity constraints,
solving the modified relaxed problem and identifying bottlenecks in its solution.
The obtained information was used to guide a proposed simulated annealing
algorithm to find a solution to the original problem.
Thus, the relaxation served only as an intermediate step toward obtaining a solution,
rather than being the desired result.

\subsection{Contribution} \label{sec:related-works/contribution}

The specific contributions of this paper are:

\begin{itemize}
    \item We extend two standard Job-Shop bottleneck identification indicators for application to the \ac{rcpsp}.
        We then develop a method utilizing the extended indicators for proposing \emph{untargeted} resource constraint relaxations.
        
    \item We develop an approach for proposing \emph{targeted} relaxations
        specifically for the \ac{rcpsp} extended with time-variant resource capacities.
\end{itemize}

\capsection{Problem statement} \label{sec:problem-statement}

\subsection{Scheduling Model} \label{sec:problem-statement/scheduling}

We assume the \Problem{} variant of the \ac{rcpsp} with several extensions to model the addressed problem.

We define a \emph{problem instance} $\Instance$ as a 4-tuple $(\Jobs, \Precedences, \Resources, \horizon)$, where
$\Jobs = \{1, \dots, n\}$ is the set of \emph{jobs},
$\Precedences$ is the set of all \emph{precedences} constraints,
$\Resources = \{1, \dots, m\}$ is the set of \emph{resources},
$\horizon\in\mathbb{N}$ is the time horizon of the problem instance.

Each job $j \in \Jobs$ has a processing \emph{duration}~$\duration{j}$ and a \emph{due date}~$\deadline{j}$.
A \emph{tardiness weight}~$\tardinessweight{j}$ defines the penalty for each time period the job is \emph{tardy},
i.e., not completed before its due date.
\emph{Preemption} of jobs is not allowed.
The order of jobs is constrained with \emph{precedence constraints}
$\precedence{i}{j}$ or $(i, j) \in \Precedences$.
We define the \emph{precedence graph} $G = (\Jobs, \Precedences)$,
which is assumed to be an \emph{inforest}, consisting of a set of connected in-trees.

Jobs are assigned to \emph{resources} $\Resources$ with time-variant renewable \emph{capacities}.
The capacity of a resource $k\in\Resources$ during a time period $t \in \intinterval{1}{\horizon}$ is denoted as $\capacity{k}{t}$.
We assume the capacities of resources to represent the availability of workers operating the resource.
Such capacities can (to some extent) be altered in correspondence to changing the number of operating workers.
For a job $j$, the per-period consumption of a resource $k$ is denoted as $\consumption{j}{k}$.
We assume that jobs can simultaneously consume multiple resources.
The resource capacity functions $\capacity{k}{t}$ are assumed to be periodic with a period of 24.
With this, we model \emph{working shifts} for the operating workers (e.g., one, two, or three-shift operations).

The set of \emph{projects} $\Projects = \{ j \in \Jobs \mid \nexists i : \precedence{j}{i} \}$
is the set of roots of the precedence in-trees.
A job $j \in \Projects$ is called a \emph{project}.
For a job $j$, due date $\deadline{j} \in \Nzero$ is given if $j \in \Projects$; $+\infty$ otherwise,
and tardiness weight $\tardinessweight{j} \geq 0$ is specified if $j \in \Projects$; $0$ otherwise, i.e., the tardiness penalty and its weight is applied to the last job of each connected component representing a single project.

\subsection{Constraint Programming Formulation} \label{sec:problem-statement/constraint-programming-model}

The above scheduling problem can be stated as the following constraint programming model:

\vspace{-1em}
\begin{align}
    \text{min}
        && \sum_{j \in \Jobs} \tardinessweight{j} \tardiness{j} \hspace{-0.5em} \label{csp:objective}
        &
        && \\
    \text{s.t.}
        && \jobend{i}
        & \leq \jobstart{j}
        & \forall \precedence{i}{j} \in \Precedences
        \label{csp:precedences} \\
    && \sum_{j \in \Jobs} \modelConsumption{j}{k}{t}
        & \leq \capacity{k}{t}
        & \hspace{-0.8em} \forall t \in \intinterval{1}{\horizon} \; \forall k \in \Resources
        \label{csp:capacities}\\
     \text{where}
        && \multicolumn{4}{l}{$\Schedule = (\jobstart{1}, \dots, \jobstart{n}) \in \N_0^n$} \nonumber\\
        && \multicolumn{4}{l}{$\Completions = (\jobstart{1} + \duration{1}, \dots, \jobstart{n} + \duration{n})$} \nonumber\\
        && \multicolumn{4}{l}{$\tardiness{j} = \max(0, \jobend{j} - \deadline{j})$} \nonumber\\
        && \multicolumn{4}{l}{$\modelConsumption{j}{k}{t} = \consumption{j}{k} \text{ if } \jobstart{j} \leq t < \jobend{j}\text{; 0 otherwise}$}\nonumber
\end{align}

The expression \eqref{csp:objective} is the optimization minimization objective --- the weighted tardiness of jobs.
Equation~\eqref{csp:precedences} formulates the precedence constraints. and Equation~\eqref{csp:capacities} describes the resource capacity constraints --- in every time period, the combined consumption of jobs scheduled during the period cannot exceed any of the resource's capacities.

We assume we have access to a solver capable of solving \eqref{csp:objective}--\eqref{csp:capacities} in a reasonable time (i.e., to provide computer-assisted decision-making to production planners)
through the use of constraint programming solvers, such as CP-SAT or IBM CP Optimizer.

\subsection{Constraints Relaxation} \label{subsec:problem-statement/constraints-relaxation}

Some constraints,
such as job precedences, job durations, or resource consumption,
are inherent to the problem (i.e., defining technological processes, physical constraints, etc.) and cannot be relaxed.
The available capacities of the resources can be modified when they reflect, e.g., the number of the available workforce at the specific stage of the production process.

We consider resource \emph{capacity additions} and \emph{capacity migrations}
as the possible relaxations of scheduling constraints \eqref{csp:capacities}.
Capacity addition is a 4-tuple $\addition{k}{s}{e}{c}$, where
over the time periods $\intinterval{s}{e-1}$ the capacity of the resource $k$ is increased by $c$.
Analogously, capacity migration is a 5-tuple $\migration{k_{\text{from}}}{k_{\text{to}}}{s}{e}{c}$, where
over the time periods $\intinterval{s}{e-1}$ the capacity of the resource $k_{\text{from}}$ is lowered by $c$
and the capacity of the resource $k_{\text{to}}$ is increased by the same amount $c$.
For a modified instance $\Instance^*$, the sets of all migrations and additions are denoted as
$\Migrations^{\Instance^*}$ and $\Additions^{\Instance^*}$, respectively.

In a real-world production system,
migrating capacities can be more cost-effective than adding new capacities.
For example, reassigning workers from an underutilized machine to a bottleneck machine is typically less expensive than extending workers' shifts into overtime or planning an entirely new and irregular shift.
Therefore, capacity migrations are usually preferred.
However, if the required capacity adjustments cannot be achieved through capacity migrations, capacity additions can be utilized.

\subsection{General Procedure} \label{subsec:problem-statement/general-procedure}

The general procedure for solving the presented problem,
identifying bottlenecks, relaxing corresponding constraints,
and solving the modified problem instance works as follows:

\begin{enumerate}
    \item Obtain a solution $\Schedule$ to the problem instance $\Instance$.
    \item Select a target project $\targetProject \in \Projects$ for tardiness improvement.
        We consider an improvement to be any non-zero decrease in the project's tardiness.
    \item Identify bottlenecks in the solution $\Schedule$ to $\Instance$.
    \item Relax constraints corresponding to the identified bottlenecks. 
        To do so, we utilize \emph{capacity migrations} and \emph{capacity additions},
        as described in \cref{subsec:problem-statement/constraints-relaxation}.
        Such relaxations are captured in a modified problem instance $\Instance^*$.
    \item Find solution $\Schedule^*$ to $\Instance^*$.
    \item Evaluate the obtained solution $\Schedule^*$.
        Specifically, how the introduced relaxations improve the tardiness of the target project.
\end{enumerate}

\capsection{Solution procedure} \label{sec:solution-apporach}

We present two algorithms designed for identifying and relaxing bottlenecks in the \ac{rcpsp}.
Both algorithms aim to improve the tardiness of a selected project
by introducing relaxations to the capacity constraints.
We divide the approaches into two groups --- untargeted and targeted relaxations depending on whether they consider the target project when identifying bottlenecks.
Thus, untargeted relaxations affect the selected project only indirectly.

\subsection{Untargeted Relaxations} \label{sec:solution-apporach/untargeted-relaxations}

In this section, we propose adaptations of existing bottleneck identification indicators
from the Job-Shop literature.
Utilizing the adapted indicators, we propose the \ac{iira}
for untargeted relaxations of capacity constraints in an obtained schedule.

\subsubsection{Adapted Identification Indicators}

We adapt two existing bottleneck identification indicators.
The \ac{mur}, first utilized as a bottleneck identification indicator by \citet{Lawrence1994},
considers the ratio of executed work on a resource to the total time the resource was used.
The \ac{auad}, initially proposed by \citet{Roser2001},
computes the average length of uninterrupted execution periods,
where an uninterrupted execution period is a sequence of scheduled immediately consecutive jobs.

Both identification indicators consider the relationship between the total duration
of job executions on a resource and the duration for which the resource is idle.
In a Job-Shop scheduling problem,
this represents all the available information.
In the \ac{rcpsp}, however,
we can utilize different resource capacities and variable resource loads
for computing more complex identification indicators.

We propose \acf{mrur} as the adaptation of \ac{mur} and \acf{auau} as the adaptation of \ac{auad}.
For resource $k$, the \ac{mrur} is defined as:
$$
\indMRUR{k} \defeq \frac{\sum_{j \in \Jobs} (\duration{j} \cdot \consumption{j}{k})}%
                        {\sum_{t=1}^{C_{\max}} \capacity{k}{t}},
$$
where $C_{\max} \defeq \max_{j \in \Jobs} \jobend{j}$.
For resource $k$, the \ac{auau} is defined as:
$$
\indAUAU{k} \defeq \frac{\sum_{i=1}^{A_k} \indPRU{k}{i}}%
                        {A_k},
$$
where the \acf{pru} $\indPRU{k}{i}$ of resource $k$ during the uninterrupted active period $i$
is defined as
$$
\indPRU{k}{i} \defeq \frac{\sum_{j \in \JobsOnResourceInPeriod{k}{i}}
                                \duration{j} \cdot \consumption{j}{k}}%
                          {\sum_{t=a_{k,i}^{S}}^{a_{k,i}^{E}} \capacity{k}{t} }.
$$
For resource $k\in \Resources$,
$(a_{k,1}^{S}, a_{k,1}^{E}), \dots, (a_{k,A_k}^{S}, a_{k,A_k}^{E})$
is the sequence of \emph{uninterrupted active periods},
where $a_{k,i}^{S} \in \intinterval{1}{\horizon-1}$ denotes the start of the period $i$
and $a_{k,i}^{E} \in \intinterval{a_{ki}^{S}+1}{\horizon}$ denotes the end of the period $i$.
An uninterrupted active period is a (maximal) set of jobs scheduled consecutively or in parallel
with no idle time on the considered resource during the period.
In the formula for $\indPRU{k}{i}$, 
$$\JobsOnResourceInPeriod{k}{i} \defeq \{ j \in \JobsOnResource{k} : a_{k,i}^{S} \leq \jobstart{j} \leq a_{k,i}^{E} \}$$
is the set of jobs executed on resource $k$ during the uninterrupted active period $i$,
where
$$
\JobsOnResource{k} = \{ j \in \Jobs : \consumption{j}{k} > 0 \}.
$$

\subsubsection{Identification Indicator-Based Relaxing Algorithm} \label{subsec:solution-approach/untargeted-relaxations/iira}

\begin{figure*}
    \centering
    \includegraphics[width=0.87\textwidth]{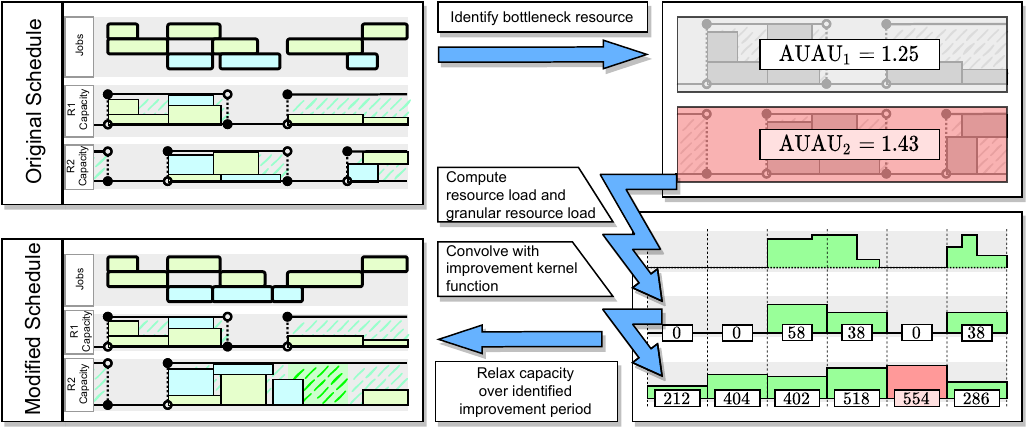}
    \caption{
        Illustration of the \acs{iira}.
        Starting with the original schedule,
        the bottleneck resource is identified using the identification indicator,
        granular resource load is computed for the resource,
        utilizing convolution, a specific improvement period is chosen for capacity relaxation,
        and a modified schedule is obtained.
    }
    \label{fig:iira}
\end{figure*}

The function of \acf{iira} is illustrated in \cref{fig:iira}
and is formally described in \cref{alg:identification-indicator-relaxing-algorithm}.

First, the bottleneck resource is identified using a specified bottleneck identification indicator
$\algIndicator \in \{ \indAUAU{}, \indMRUR{} \}$
(lines~\ref{alg:iira/evaluation} and~\ref{alg:iira/identification}).
The granular load of the bottleneck resource (line~\ref{alg:iira/granular-load})
indicates the resource's utilization over granular periods --- granular periods each represent $\algGranularity$ time periods for efficient computation.
Convolution with a chosen kernel function is used to obtain the improvement potential of granular periods
(line~\ref{alg:iira/convolution})
and periods with the most improvement potential are then selected for capacity increase,
relaxing the represented capacity constraints
(lines~\ref{alg:iira/periods} and \ref{alg:iira/capacity-increase}).
The convolution \enquote{distributes} the local information about the machine load
to adjacent periods to estimate which periods to focus on.
Finally, a new solution to the modified problem instance is found (line~\ref{alg:iira/modified-solution}),
capacity functions are reduced to only contain additional capacities consumed by jobs,
and migrations and additions are identified.

\begin{algorithm}[t]
\caption{\acf{iira}}
\label{alg:identification-indicator-relaxing-algorithm}
\begin{algorithmic}[1]
    
\Params  Identification indicator $\algIndicator$, convolution kernel $\algConvolution$, granularity $\algGranularity$, %
    improvement periods limit $\algMaxperiods$, iterations limit $\algMaxiter$, %
    capacity improvement $\algImprovement$.
\Input  Solution $\Schedule$ to problem instance $\Instance$.

\Repeat \label{alg:iira/repeat}
    \State Evaluate $\Schedule^*$ using $\algIndicator$, obtaining:
           $\algIndicator_k \;\forall k \in \Resources$ \label{alg:iira/evaluation}
    \State Identify bottleneck resource:
           $k^\prime \gets \argmax_k \algIndicator_k$\label{alg:iira/identification}
    \State Compute granular resource load $\resourceLoad{k^\prime}(G)$\label{alg:iira/granular-load}
    \State Period improvement potential: $\Psi \gets \resourceLoad{k^\prime}(G) * \algConvolution$\label{alg:iira/convolution}
    \State Select periods with highest $\Psi(i)$: $p_1, \dots, p_{\algMaxperiods}$\label{alg:iira/periods}
    \State Increase capacity $\capacityf{k^\prime}^\prime$ over the periods by $\algImprovement$ \label{alg:iira/capacity-increase}
    \State Find solution $\Schedule^*$ to the modified instance $\Instance^*$ \label{alg:iira/modified-solution}
    \State Reduce capacity changes in $\capacityf{1}^\prime, \dots, \capacityf{m}^\prime$
    \State Find migrations $\Migrations^{\Instance^*}$ and additions $\Additions^{\Instance^*}$
\ForIter{$\algMaxiter$} \label{alg:iira/for-iters}

\Output  Modified instance $\Instance^*$ and its solution $\Schedule^*$,
\Outputc additions $\Additions^{\Instance^*}$, migrations $\Migrations^{\Instance^*}$.

\end{algorithmic}
\end{algorithm}

\subsection{Targeted Relaxations} \label{sec:solution-approach/targeted-relaxations}

As an alternative to untargeted relaxations, in this section, we present a method for detecting bottlenecks and relaxing related constraints in the \ac{rcpsp}
which focuses on a specified target project and its tardiness.
The proposed  \acl{ssira}  is based on finding improvement intervals
in partially relaxed versions of the given problem.
A small subset of the improvement intervals is then selected, and capacity constraints corresponding to the selected improvement intervals are relaxed.
The targeted relaxation aims to identify relaxations specifically for the target project requiring small changes to improve the tardiness of the project.

\subsubsection{Preliminaries} \label{subsec:solution-approach/targeted-relaxations/preliminaries}

To formulate \acf{ssira}, we state the necessary definitions and the key ideas.
First, we define the \emph{suffix-relaxed schedule}, which is a modification of an obtained schedule where the algorithm finds improvement intervals.

\begin{defn}[Suffix-relaxed schedule] \label{def:suffix-relaxed-schedule}
    Let $\Schedule = (\jobstart{1}, \dots, \jobstart{n})$ be a schedule to a problem instance $\Instance$.
    Given a time period $t \in \intinterval{1}{\horizon}$,
    the \emph{suffix-relaxed schedule} for the time period $t$ is given by
    $\relaxedSchedule{t} = (\relaxedjobstart{t}{1}, \dots, \relaxedjobstart{t}{n})$, where
    $$
    \relaxedjobstart{t}{j} \defeq \begin{cases}
        \jobstart{j} & \text{if $\jobstart{j} \leq t$;} \\
        \max\left\{\relaxedjobstart{t}{i} + \duration{i}: \precedence{i}{j} \in \Precedences \right\} & \text{otherwise.}
    \end{cases}
    $$
\end{defn}

The precedence graph is acyclic;
thus, all values of $\relaxedjobstart{t}{i}$ are well-defined.
The suffix-relaxed schedule essentially relaxes resource capacity constraints for all jobs that start after the time period $t$ in the original schedule.
The main idea is to incrementally observe how jobs that are constrained by insufficient capacities and precedence constraints could be scheduled earlier.

We define the \emph{left-shift closure} as a tool
for guiding the search for improvement intervals towards improving the tardiness of the target project.
The left-shift closure of a job $j$ defines the set of all jobs
that need to be scheduled earlier for the job $j$ to decrease its completion time.

\begin{defn}[Left-shift closure] \label{def:left-shift-closure}
    Let $\Schedule = (\jobstart{1}, \dots, \jobstart{n})$ be a schedule to problem instance $\Instance$.
    The \emph{left-shift closure} of a job $j \in \Jobs$ is the set $\closure{j}~\subseteq~\Jobs$, where:
    \begin{conditions}
        \item
            $j \in \closure{j}$.
            \label{def:closure/base}

        \item
            All precedence predecessors immediately preceding in the schedule are included.
            \label{def:closure/precedence}

        \item
            All jobs sharing a common resource immediately preceding in the schedule are included.
            \label{def:closure/resource-precedence}

        \item
            If $j$ is scheduled exactly at any start of a resource's availability interval,
            all jobs scheduled at the end of the previous availability interval are included.
            For a given resource, an availability interval is a sequence of consequtive time periods
            where the resource's capacity is non-zero.
            \label{def:closure/shift-pause-precedence}
    \end{conditions}
\end{defn}

\Cref{def:closure/base} is a trivial base case.
\Cref{def:closure/precedence} states that a immediate precedence predecessor $i$ of the job $j$
(i.e., $\precedence{i}{j} \in \Precedences$) is included in $\closure{j}$ if the jobs are scheduled consecutively,
i.e. $\jobend{i} = \jobstart{j}$.
\Cref{def:closure/resource-precedence} involves all jobs scheduled consecutively before the job $j$,
which share at least one required resource.
In this case, the consumption of the shared resource by the preceding job can be sufficiently large to prevent the job $j$ from being scheduled earlier.
\Cref{def:closure/shift-pause-precedence} involves jobs at the end of the previous working shift.
Assuming sufficient slack in precedence constraints,
the job $j$ starts exactly at the start of a working shift because
it could not have been scheduled at the end of the previous working shift due to the lack of remaining capacities on its required resources.

\subsubsection{Schedule Suffix Interval Relaxing Algorithm} \label{subsec:solution-approach/targeted-relaxations/ssira}

\begin{figure*}
    \centering
    \includegraphics[width=0.87\textwidth]{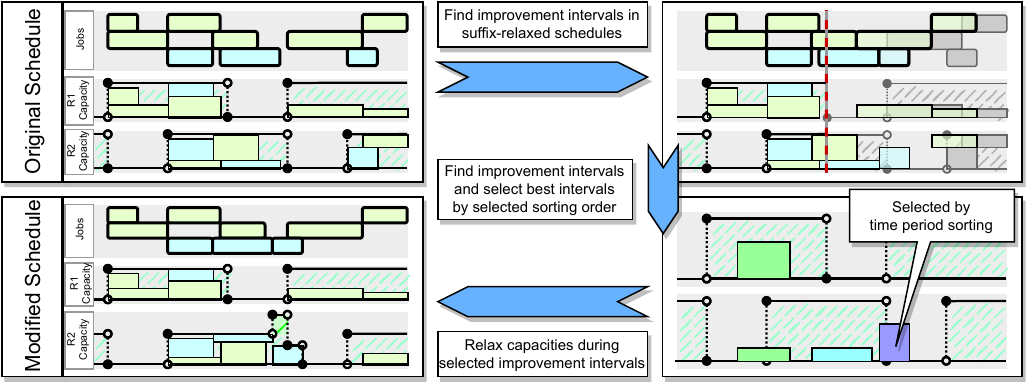}
    \caption{
        Illustration of the \acs{ssira}.
        Starting with the original schedule,
        improvement intervals are found in suffix-relaxed schedules,
        the best improvement intervals are selected, corresponding capacities are relaxed,
        and a new schedule is obtained.
    }
    \label{fig:ssira}
\end{figure*}

The schematic highlight of  \ac{ssira} is displayed in \cref{fig:ssira}
and is formally given in \cref{alg:schedule-suffix-interval-relaxing-algorithm}.
First, improvement intervals are identified using the \algnameref{FindIntervalsToRelax}{alg:find-intervals-to-relax} function
(line~\ref{alg:ssira/ints}).
Resource capacities are then relaxed based on these intervals (line~\ref{alg:ssira/modify}).
Finally, as in the \ac{iira},
a solution to the modified instance is found,
capacity functions are reduced,
and migrations and additions are identified.

The \algnameref{FindIntervalsToRelax}{alg:find-intervals-to-relax} function,
described in \cref{alg:find-intervals-to-relax},
finds improvement intervals in suffix-relaxed schedules and selects the best intervals based on a given ordering.
Suffix-relaxed schedules are computed for each time period (line~\ref{alg:ssira/ints/schedule-suffixes}),
representing all possible job-interval relaxations.
The left-shift closure of the target project is computed (line~\ref{alg:ssira/ints/closure}).
This closure represents the set of jobs considered for improvement.
For jobs within the closure,
potential improvement intervals are identified (lines~\ref{alg:ssira/ints/ints-init}--\ref{alg:ssira/ints/ints-inclusion}),
where the starting time of each interval is the earliest relaxed starting time
across all suffix-relaxed schedules (line~\ref{alg:ssira/ints/ints-improvement}).
Finally, a given number of intervals is selected based on a specified sort key (line~\ref{alg:ssira/ints/select-imp-ints}).

\begin{algorithm}
\caption{\acf{ssira}}
\label{alg:schedule-suffix-interval-relaxing-algorithm}
\begin{algorithmic}[1]

\Params  Iterations limit $\algMaxiter$, improvement intervals limit $\algMaxintervals$, interval sort key $\algSortkey$.
\Input  Solution $\Schedule$ to instance $\Instance$,
target project $\targetProject$.
\State $\Instance^* \gets \Instance$, $\Schedule^* \gets \Schedule$
       \Comment Modified instance and solution
\Repeat \label{alg:ssira/repeat}
    \State $\chi_1, \dots, \chi_{\algMaxintervals} \gets$ \Callref{FindIntervalsToRelax}%
                                                                  {}{alg:find-intervals-to-relax}
                                                                  \label{alg:ssira/ints}
    \State Increase capacities $\capacityf{1}^*, \dots, \capacityf{m}^*$ in the intervals
                                                                   \label{alg:ssira/modify}
    \State Find solution $\Schedule^*$ to the modified instance $\Instance^*$ \label{alg:ssira/solution}
    \State Reduce capacity changes in $\capacityf{1}^*, \dots, \capacityf{m}^*$ \label{alg:ssira/reduction}
    \State Find migrations $\Migrations^{\Instance^*}$ and additions $\Additions^{\Instance^*}$ \label{alg:ssira/additions-migrations}
\ForIter{$\algMaxiter$} \label{alg:ssira/for-iters}

\Output  Modified instance $\Instance^*$ and its solution $\Schedule^*$,
\Outputc additions $\Additions^{\Instance^*}$, migrations $\Migrations^{\Instance^*}$.
\end{algorithmic}
\end{algorithm}

\vspace{-0.5cm}
\begin{algorithm}
\caption{FindIntervalsToRelax}
\label{alg:find-intervals-to-relax}
\begin{algorithmic}[1]

\Input  Problem instance $\Instance$, its solution $\Schedule$, improvement intervals limit $\algMaxintervals$,
        interval sort key $\algSortkey$, target project $\targetProject$.

\State Compute suffix-relaxed schedules $\relaxedSchedule{1}, \ldots, \relaxedSchedule{\horizon}$ \label{alg:ssira/ints/schedule-suffixes}
\State Compute left-shift closure $\closure{\targetProject}$ \label{alg:ssira/ints/closure}
\State $X \gets \emptyset$  \label{alg:ssira/ints/ints-init}
\For {$j \in \closure{\targetProject}$}
    \State $s \gets \min_t \left\{ \relaxedjobstart{t}{j} : \relaxedjobstart{t}{j} < \jobstart{j} \right\}$  \label{alg:ssira/ints/ints-improvement}
    \State $X \gets X \cup \left\{ (j,\; s,\; s + \duration{j}) \right\}$  \label{alg:ssira/ints/ints-inclusion}
\EndFor

\State Find first $\chi_1, \dots, \chi_{\algMaxintervals}$ from $X$ ordered by $\algSortkey$ \label{alg:ssira/ints/select-imp-ints}

\Output  Improvement intervals $\chi_1, \dots, \chi_{\algMaxintervals}$,
\Outputc a set of 3-tuples $(j, s, e) \in \Jobs \times \intinterval{1}{\horizon}^2$.
\end{algorithmic}
\end{algorithm}

\capsection{Experiments} \label{sec:numerical-experiments}

We evaluate the performance of an untargeted bottleneck detection method called \acf{iira} and a targeted method called \acf{ssira}.
We first design benchmark instances that model the addressed problem
and choose ranges of parameters for each algorithm,
creating evaluation parameter sets.
Then, we conduct the experiments, make several observations about the outcomes,
and discuss the achieved results.

\subsection{Setup} \label{sec:numerical-experiments/setup}

We use and modify specific instances from the PSPLIB single-mode instance set.
The modifications include 
splitting the precedence graph to create individual project components,
introducing job due dates,
introducing time-variable resource capacities,
and scaling down job durations and resource consumptions for otherwise infeasible instances.
We propose eight problem instance groups,
each consisting of five individual instances of similar properties
(e.g., precedence graph structure or project due dates).

We use the IBM \acs{cpopt} for finding optimal solutions.
The solver time limit for finding a single solution was set to 10 seconds.
Subsequent solving of modified instances utilizes solver warm-starting.
The experiments and the created instances can be found on \href{https://github.com/Krtiiik/RCPSPSandbox}{GitHub}\footnote{
\url{https://github.com/Krtiiik/RCPSPSandbox}
}.

For both algorithms \acs{iira} and \acs{ssira}, all combinations of parameter values are considered, forming a total of 288 combinations for the \ac{iira} and 36 combinations for the \ac{ssira}.
Algorithms are evaluated with each combination of values on every problem instance.

The following metrics are computed:
\begin{itemize}
    \item
        Tardiness improvement
        $
        \kpiImprovement \defeq \tardiness{\targetProject} - \tardiness{\targetProject}^*.
        $
        This metric is also meaningful for the \ac{iira} (i.e., untargeted), where we measure $\kpiImprovement$ for the target project $\targetProject$.
    
    \item
        Solution difference
        $
        \kpiDiff \defeq \sum_{j \in J} \abs{\jobend{j} - \jobend{j}^*}.
        $
\end{itemize}

\subsection{Comparative Results} \label{sec:numerical-experiments/comparative-results}


\begin{table}
    \centering
    \caption{
        Improving solutions found for the proposed problem instances.
        The \emph{Improved} value states the number of instances the algorithm found an improving solution for,
        the \emph{Unique} value states how many were uniquely found w.r.t. the other algorithm,
        and the \emph{Best} value states how many were the best-improving solutions.
    }
    \begin{tabular}{llcc}
        \toprule
        Algorithm & Criteria & Solutions & \% (of 40) \\
        \midrule
             & Improving & $29$ & $\pmb{72.5\%}$ \\
        IIRA & Unique & $0$ & $0\%$ \\
             & Best & $22$ & $55\%$ \\
        \midrule
              & Improving & $35$ & $\pmb{87.5\%}$ \\
        SSIRA & Unique & $6$ & $15\%$ \\
              & Best & $25$ & $62.5\%$ \\
        \bottomrule
    \end{tabular}
    \label{tab:exp/improvements}
\end{table}

\Cref{tab:exp/improvements} summarizes achieved improvements.
The \ac{ssira} found improvements for more instances than the \ac{iira},
moreover, \ac{iira} did not improve any instance which the \ac{ssira} would not improve.

\begin{figure}
    \centering
    \includegraphics[width=\linewidth]{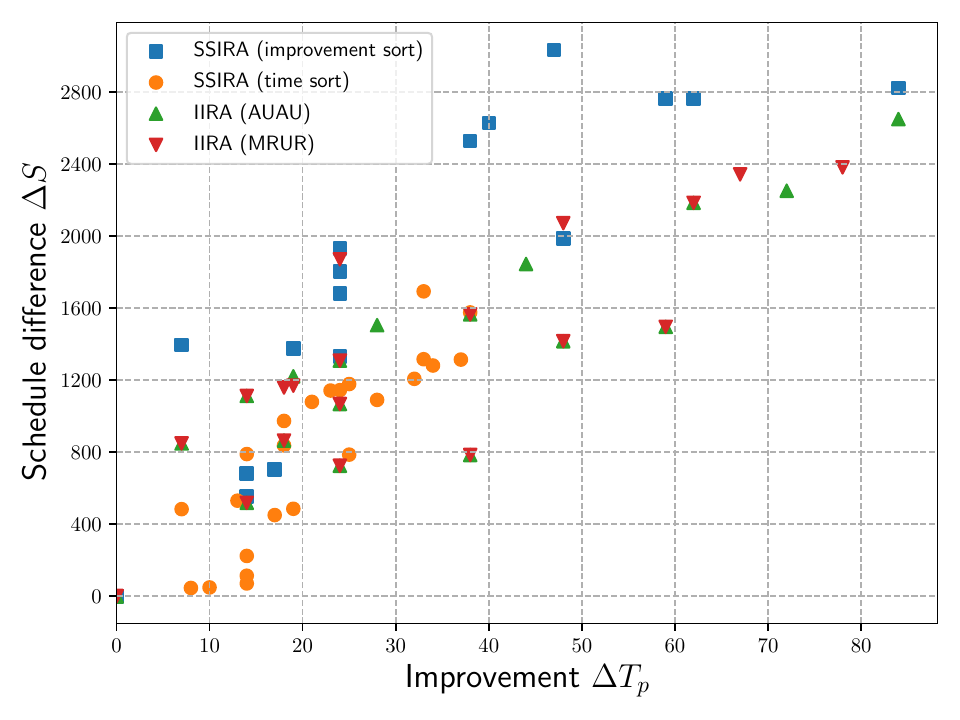}
    \caption{
        An example evaluation displays schedule differences versus achieved improvement. Surprisingly, the \ac{iira} often finds better solutions than \ac{ssira}.
        }
    \label{fig:exp/improv-diff}
\end{figure}

As expected, greater tardiness improvements generally induce larger schedule differences.
The \ac{ssira} utilizing the $\algSortkeyImprovement$ sort key tends to propose
the least favorable solutions in terms of the induced schedule difference
and is the most inconsistent in finding improving solutions.
The \ac{ssira} with the $\algSortkeyTime$ sort key finds improving solutions consistently
across many instances.
However, for some instances, the \ac{iira} is able to find better solutions than the \ac{ssira}.
In \cref{fig:exp/improv-diff}, we present an example of results concerning
tardiness improvement related to induced schedule difference
showcasing the aforementioned trends.

\subsection{Discussion} \label{sec:numerical-experiments/discussion}

The \ac{ssira} finds an improvement more often than the \ac{iira}.
We believe this is because the \ac{ssira}, unlike the \ac{iira}, utilizes targeted relaxations
focusing on the target project.
However, the \ac{iira} was still able to find many improving solutions, sometimes even
surpassing the performance of the targeted relaxations proposed by the \ac{ssira}.
This is an unexpected result,
as the initial assumption was that targeted relaxations
would achieve better improvements than general relaxations.
It seems to us that targeted relaxations of the \ac{ssira} might be too specific, not providing sufficient slack in the modified constraints
and thus making the model too sensitive to minor variations when finding solutions for the relaxed problem.
In addition, focusing only on the jobs from the left-shift closure of the target project might be a good heuristic,
but it might be too restrictive.
Another possibility is that the \ac{ssira} often proposes multiple relaxations simultaneously,
incorrectly assuming their independence.

\capsection{Conclusion} \label{sec:conslussion}

We addressed the problem of bottleneck identification in production schedules as a computer-aided tool for production planners.
First, we formulated an extension of the standard \ac{rcpsp} as a simplified model of production.
Then, we focused on execution-level machine bottlenecks in obtained schedules.
Following the identification of such bottlenecks,
we proposed constraint relaxations for related resource capacity constraints to find a solution of better quality.

We extended two well-known Job-Shop bottleneck identification indicators for the \ac{rcpsp}.
We proposed the \ac{iira}, utilizing the extended indicators and untargeted relaxations.
We also proposed the \ac{ssira}, designed to utilize targeted relaxations.

We observed that the \ac{ssira} is more consistent in finding improving solutions than the \ac{iira}.
However, for many instances, the \ac{iira} is able to find great improvements with lower induced schedule differences than those proposed by the \ac{ssira}.
Thus, untargeted  methods utilizing bottleneck identification indicators
appear to be promising in the \ac{rcpsp},
even for a specific (targeted) project.

Future work might involve modeling the relaxations as an optimization problem
to better capture the complex dependencies of the considered constraints,
or further exploring the use of bottleneck identification indicators in the \ac{rcpsp} and its applications in related problems such as 3-dimensional spatial \ac{rcpsp}~\parencite{Zhang2024}.

\section*{ACKNOWLEDGEMENTS}
This work was supported by the Grant Agency of the Czech Republic under the Project GACR 22-31670S, and was co-funded by the European Union under the project ROBOPROX (reg. no. CZ.02.01.01/00/22\_008/0004590).

\renewcommand{\refname}{REFERENCES}
\printbibliography{}

\end{document}